# The Design of Autonomous UAV Prototypes for Inspecting Tunnel Construction Environment

## Yiping Dong*

Department of Mechanical Engineering, Carnegie Mellon University, Pittsburgh, PA, 15213, United States;

*Corresponding Author: dand97personal@gmail.com

## ABSTRACT

This article presents novel designs of autonomous UAV prototypes specifically developed for inspecting GPS-denied tunnel construction environments with dynamic human and robotic presence. Our UAVs integrate advanced sensor suites and robust motion planning algorithms to autonomously navigate and explore these complex environments. We validated our approach through comprehensive simulation experiments in PX4 Gazebo and Airsim Unreal Engine 4 environments. Real-world wind tests and exploration experiments demonstrate the UAVs' capability to operate stably under diverse environmental conditions without GPS assistance. This study highlights the practicality and resilience of our UAV prototypes in real-world applications.

Keywords: Autonomous UAV, Tunnel Inspection, GPS-Denied Environment, Safe Trajectory Planning, Real-World Testing

## 1. Introduction and Background

### 1.1 Introduction

In a variety of civilian and military applications such as surveillance, inspection, search, and rescue, robotic systems are gaining importance and becoming increasingly useful. Especially, well-developed autonomous systems are always expected to keep humans from the risk of operating in a dangerous and unknown environment. However, environments with such characteristics are usually more challenging for autonomous systems operations. For example, in GPS-denied environments, the robot is required to estimate its state and make decisions solely on sensor measurements and without access to precise position information [1]. In industrial scenarios with a complex structured environment with moving humans and robots, as shown in Fig. 1, autonomous UAVs are required to navigate to their goals in cluttered environments and guarantee safety with humans. Consequently, a stable UAV hardware platform and a safe trajectory planning software framework are essential to deal with complex environment structures, dynamic obstacles, and uncertainties from measurement noise and unpredictable moving obstacle behaviors [2].

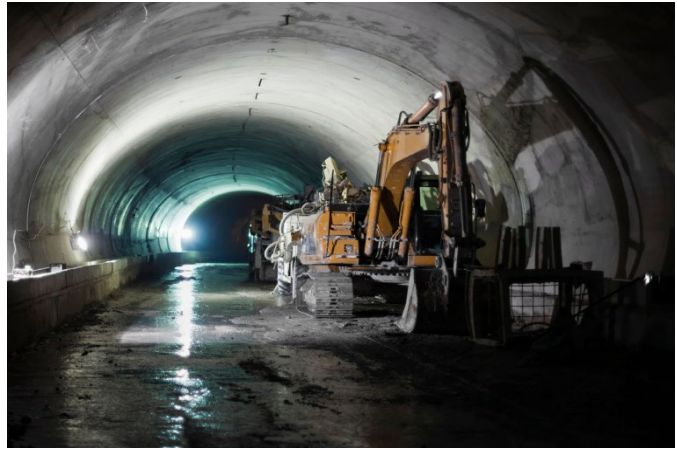

Figure 1. The complex structured environment with moving humans and robots

There are two main challenges to building a stable UAV hardware platform and a safe trajectory planning software framework. Firstly, the complexity of the human-robot collaboration environment imposes many constraints on UAV's parameters, such as the frame size, the total payload, and the sensor type. Secondly, given constraints on UAV's attributes, a safe trajectory planning framework should convert these hardware constraints into corresponding software requirements and consider them when designing and implementing the safe trajectory planning framework. Even though some works prove the exploration performance in subterranean environments [1] [3] [4], the real-time obstacle avoidance ability is not fully shown. Previous publications [5] [6] also show the mature exploration and obstacle avoidance ability under the ground mobile robots and aerial robots collaboration circumstances, the ability to use aerial robots to independently finish tasks is not shown.

## 1.2 Background

UAV hardware platform designs for GPS-denied tunnel construction environments can be customized based on specific requirements. In [7], a new type of collision-tolerant robot called resilient micro flyer is designed to do navigation in confined environments. The robot maintains a low weight (<500g), and small frame (0.32m-diameter) and implements a combined rigid-complaint design through the integration of an elastic flap around its stiff collision-tolerant frame. In [8], to provide fast exploration ability, especially in the areas not accessible by the ground robots, a team of Aerial Scouts is used. This aerial robot team is fully functional for the construction tunnel environment exploration while considering some extreme working disturbances. There are mainly three categories of these scouts, namely, medium-sized multirotor, small-sized collision-tolerant flying robots (Gagarin and RMF-Owl), and a single large-sized tricopter (Kolibri). The RMF-Owl and the Gagarin platforms integrated LiDAR and vision and were two types of collision-tolerant aerial vehicles featuring a protective outer cage. Gagarin was developed around the Elios collision-tolerant platform built by Flyability. The medium-sized scouts, built around the M100 frame by DJI, integrated a LiDAR, a color camera, and a thermal camera and could penetrate dusty and smoke-filled environments. The large-sized Kolibri, based on a tricopter platform from Voliro, was intended for long-term missions (over 20-minute endurance) and vertical exploration (by being able to pitch independently of its translation). In general, nearly all sizes of aerial robot prototypes are used in this paper to enable collaborations.

Safe trajectory planning software framework development is another topic that has been investigated recently regarding tunnel exploration and navigation problems. Normally, the planning software framework can be divided into two parts, the low-level local planning part, and the high-level autonomous exploration decision-making and path planning part. [9] proposes a local planning method for legged robots using reachability planning and template learning to navigate in rough terrain, based on their previous work [10] on PRM*, a persistent planning graph between planning queries. As the planner mentioned before covers the environments with few dynamic obstacles such that the previously observed map regions remain largely unchanged. In [10], another proposed single-query planner that is like RRT* can additionally handle the scenarios that require a fast update rate to keep up with frequent drastic map changes.

In [3], the graph-based subterranean exploration path planning algorithm using aerial and legged robots shows its ability by utilizing a rapidly exploring random graph to reliably and efficiently identify paths that optimize an exploration gain within a local subspace while simultaneously avoiding obstacles, respecting applicable traversability constraints and honoring dynamic limitation of the robots. A similar centralized robots' system and exploration strategy are also mentioned in [4] when a stable communication link is enabled. While both [3] and [4] prove to have a satisfying performance on exploration tasks in target environments, the learning-based exploration path planner in [1] uses the proposed planner in [3] and [4] as a "training expert" and following an approach relying on the concepts of imitation learning. The algorithm utilizes only a short window of range data sampled from the onboard LiDAR and achieves an exploratory behavior similar to that of the training expert with a more than an order of magnitude reduction in computation cost, while simultaneously relaxing the need to maintain a consistent and online reconstructed map of the environment.

Apart from the aerial robots and ground robots' collaboration exploration algorithms, [11] shows the exploration algorithm that only uses an aerial robot. A general problem that exists before is that during exploration, robots oftentimes must rely on onboard systems alone for state estimation, accumulating significant drift over time in a large environment. This drift can be detrimental to robot safety and exploration performance. Compared to previous work in [1] [3] [4], this algorithm solves the problem by proposing a submap-based, multi-layer approach for both mapping and planning to enable safe and efficient volumetric exploration of large-scale environments despite the odometry drift mentioned before.

In this paper, we present novel designs of UAV prototypes for inspecting tunnel construction environments. This prototype contains two main parts to deal with the designated tunnel inspection tasks in the GPS-denied environment. In the stable UAV hardware development part, we propose two different designs using separate sets of sensors and controllers. Then in the safe trajectory planning software framework development part, we propose a software pipeline that includes environment information acquisition, uncertain area exploration, collision-free path generation, and trajectory optimization. Extensive real-world experiments and simulation experiments are performed to demonstrate that our prototypes can help UAVs safely navigate in a complex structured environment with moving humans and robots. The novelties and contributions of this work are:

- Novel Designs of UAV Prototypes: This work presents two distinct UAV hardware platform designs, and both are developed under the environment and industrial requirements.
- High-Efficiency Path Planner: The proposed path planner can effectively generate a collision-free path under the limitations of the input sensors' information types and computation resources.
- Customized Tasks and Experiment Environments: The proposed customized tasks and experiment environments can help verify the performance of the UAV prototypes while preventing leading to any actual risk and fitting the real industrial requirements at the same time.

## 2. Related Work

### 2.1 Flyability Elios 2 in Tunnel Inspection Applications

Flyability Elios 2 [12] is a pioneering UAV model specifically engineered for navigating and inspecting confined and GPS-denied environments, making it highly relevant in tunnel construction inspections. This spherical, collision-tolerant drone is designed to operate safely in complex and cluttered spaces where traditional UAVs struggle to maneuver effectively. Equipped with high-definition and thermal imaging cameras, Elios 2 provides real-time visual feedback crucial for detailed inspections in environments characterized by limited visibility and potential hazards.

In tunnel inspection applications, Flyability Elios 2 excels due to its robust design and specialized capabilities [13]. The drone's spherical cage structure protects its core components, allowing it to navigate close to walls, ceilings, and obstacles without risking damage. This capability is particularly valuable in assessing structural integrity, identifying cracks, and inspecting hard-to-reach areas where manual inspection is impractical or hazardous. Elios's ability to capture high-quality images and thermal data ensures comprehensive monitoring of tunnel conditions, aiding in proactive maintenance and safety management.

Flyability Elios 2 offers several distinct advantages in tunnel inspections. Its collision-tolerant design enables safe exploration and detailed visual inspection in confined spaces, reducing the risk of equipment damage and improving operational efficiency. The integration of high-definition and thermal cameras provides precise imaging capabilities, facilitating accurate defect detection and environmental monitoring [14]. Moreover, the drone's agility and maneuverability enhance its adaptability to dynamic tunnel environments, ensuring thorough and reliable inspection outcomes.

Despite its strengths, Flyability Elios 2 has limitations primarily related to payload capacity and operational endurance. The spherical cage, while protective, adds weight and limits the size of payload that can be carried, restricting the range of additional sensors or equipment that can be integrated. Additionally, the drone's flight time is typically constrained to around 20-30 minutes per battery charge, necessitating strategic planning for longer inspection tasks or large-scale tunnel networks.

Flyability Elios 2 represents a significant advancement in UAV technology tailored for tunnel inspection applications. Its specialized design and capabilities address critical challenges in navigating GPS-denied and confined environments, providing detailed and actionable insights for infrastructure maintenance and safety. While its protective design and imaging capabilities enhance operational safety and efficiency, considerations such as payload limitations and flight endurance remain factors to optimize in future iterations. Overall, Elios 2's application in tunnel inspections showcases the potential of advanced UAVs to revolutionize infrastructure management through enhanced inspection methodologies.

## 2.2 DJI Matrice 300 RTK in Tunnel Inspection Applications

The DJI Matrice 300 RTK [15] represents a cutting-edge UAV model tailored for demanding industrial applications, including tunnel inspection where precise navigation and comprehensive data collection are essential. Equipped with Real-Time Kinematic (RTK) positioning technology, this multi-rotor platform offers unparalleled accuracy in GPS-denied environments, making it highly suitable for navigating complex tunnel structures with minimal external guidance.

In tunnel inspection scenarios, the DJI Matrice 300 RTK excels due to its robust construction and advanced sensor integration capabilities. The UAV's RTK system ensures centimeter-level positioning accuracy, crucial for detailed mapping and spatial awareness within tunnels where GPS signals are unreliable. Integrated with a suite of sensors including LiDAR, RGB cameras, and thermal imaging modules, the Matrice 300 RTK facilitates comprehensive inspections by capturing high-resolution images, 3D point clouds, and thermal data [16] [17]. These capabilities enable precise monitoring of structural integrity, identification of potential hazards such as cracks or leaks, and assessment of environmental conditions affecting tunnel safety.

The DJI Matrice 300 RTK offers several advantages in tunnel inspection applications. Its RTK positioning technology ensures precise navigation and mapping even in GPS-deprived environments, enhancing operational reliability and accuracy [18]. The integration of multiple sensors enables versatile data acquisition, from detailed visual inspections to comprehensive environmental monitoring using thermal imaging. Moreover, the UAV's extended flight time (up to 55 minutes per battery) and robust build quality enhance its suitability for long-duration missions and challenging operational conditions.

Despite its strengths, the Matrice 300 RTK has limitations primarily related to complexity and cost. The integration of advanced sensors and RTK technology increases the initial investment and operational complexity compared to simpler UAV models. Furthermore, while its flight endurance is notable, it may still require careful planning for extended missions or large-scale tunnel networks to manage battery life effectively. Additionally, the UAV's size and weight can pose challenges in confined spaces or narrow tunnel sections, potentially limiting its maneuverability in certain operational scenarios.

The DJI Matrice 300 RTK represents a significant advancement in UAV technology for tunnel inspection applications, leveraging precise navigation, advanced sensor capabilities, and extended flight endurance to enhance infrastructure management and safety. Its ability to deliver high-quality data and perform complex inspections in challenging environments underscores its value in modern infrastructure inspection practices. While considerations such as cost, complexity, and operational constraints exist, the Matrice 300 RTK remains a powerful tool for professionals seeking reliable and comprehensive solutions in tunnel construction and maintenance.

## 3. Experimental / Computational Methods

### 3.1 System Framework

The system for exploration experiments mainly has four parts: a laptop computer with Ubuntu 18.04 and ROS Melodic, a transmitter, any mobile device compatible with QGroundControl, and the drone mounted with cameras and an onboard computer as shown in Fig. 2. To enable the experiments can be conducted safely and efficiently, the laptop computer gets the odometry information about the drone from onboard computer to monitor if the status of the drone is correct before and during the experiment. Then the mobile device will use QGroundControl to get battery information, current flight mode, and error messages while testing. The transmitter helps send commands from the human manipulator to the drone to end a failed experiment timely to prevent any crash.

The system components and interactions can be described using mathematical notation:

Odometry Information Monitoring:

$$\text{Odometry Information: } x_t, y_t, z_t \qquad \text{[Formula 1]}$$

The laptop computer monitors the odometry information $(x_t, y_t, z_t)$ from the onboard computer to ensure correct drone status during experiments.

Command Transmission:

$$\text{Command: } u(t) = \big(u_1(t), u_2(t), u_3(t)\big) \qquad \text{[Formula 2]}$$

Commands $(u(t))$ are transmitted from the human manipulator via the transmitter to control the drone's actions and prevent crashes during failed experiments.

State Estimation Fusion:

$$\text{Fusion of Odometry and Localization: } \hat{x}_t = f(x_t, z_t) \qquad \text{[Formula 3]}$$

The fusion process $(\hat{x}_t = f(x_t, z_t))$ integrates odometry $(x_t, y_t, z_t)$ and localization $(z_t)$ to estimate the drone's state accurately.

### 3.2 Customized Tasks and Experiment Environments

Given a general exploration task in the target complex structured environment with moving humans and robots, there are always multiple possible ways to divide them into several customized subtasks and construct corresponding simulations and real-world test fields to validify the proposed UAV prototypes.

Based on our general exploration task, we propose in total four subtasks for our UAV prototypes to complete: take off at the entrance of the tunnel, navigate to the dead-end of the tunnel while constructing the map, and make the dynamic and static obstacle avoidance at the same time, make three-dimensional reconstruction for the dead-end of the tunnel, navigate back to the entrance using the constructed map.

For the experiment environments, there are mainly three types of experiment setups in this paper. Firstly, we set up the simulation environments to separately test the functionalities of our high-efficiency path planner's ability to finish the mentioned subtasks. Secondly, the real-world exploration test to validify the robustness and accuracy of our designed UAV models in executing the commands sent from our planner. Thirdly, a supplemental wind test is designed to make sure that the controller can automatically calibrate the drift and keep the UAV at the target position. We use a fan with three different power levels to create wind fields for the drone. The UAV's drifts under a hovering state and going straight state while the wind flows from the side of the UAV are evaluated and compared with the performance without any wind.

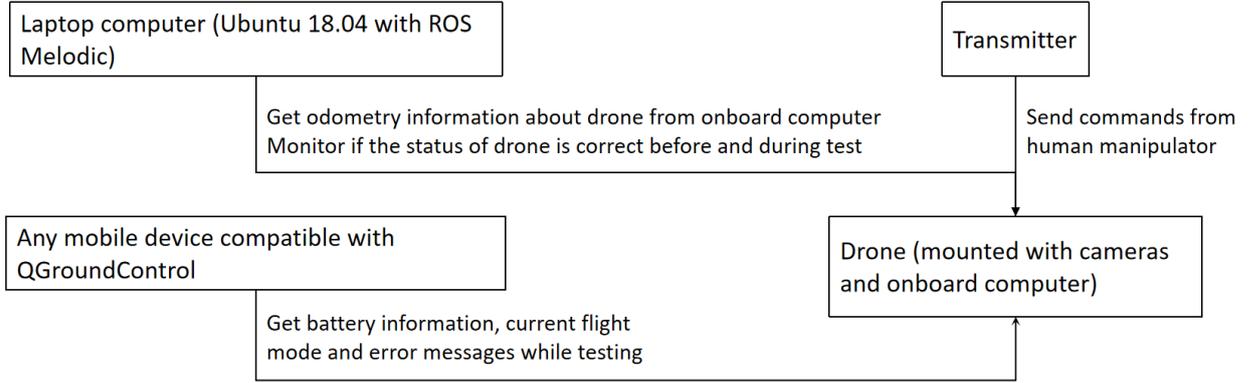

Figure 2. System overview. The figure shows four essential components required for the real-world exploration experiments.

Describing the task divisions and experiment setups with mathematical clarity:

Task Segmentation:

$$\text{Tasks: } T_1, T_2, \ldots, T_n \qquad \text{[Formula 4]}$$

Tasks $(T_1, T_2, \ldots, T_n)$ are segmented to validate UAV prototypes in various scenarios within complex structured environments.

Simulation Setup:

$$\text{Simulation Environment: } SE = \{E_1, E_2, \ldots, E_m\} \qquad \text{[Formula 5]}$$

Simulation environments $(SE)$ are configured to assess the performance of the path planner in diverse scenarios $(E_1, E_2, \ldots, E_m)$.

Wind Test Evaluation:

$$\text{Wind Effect Evaluation: } W = \{W_1, W_2, W_3\} \qquad \text{[Formula 6]}$$

Wind tests $(W = \{W_1, W_2, W_3\})$ validate the UAV's ability to maintain stability under varying wind conditions.

## 3.3 Novel Designs of UAV Prototypes

There are two different designs of UAV prototypes. Both UAV prototypes can finish the designed tasks for exploration in tunnel construction environments. Based on the time sequence two UAV prototypes are designed, we name them UAV prototype version one, as shown in Fig. 3, and UAV prototype version two, as shown in Fig. 4.

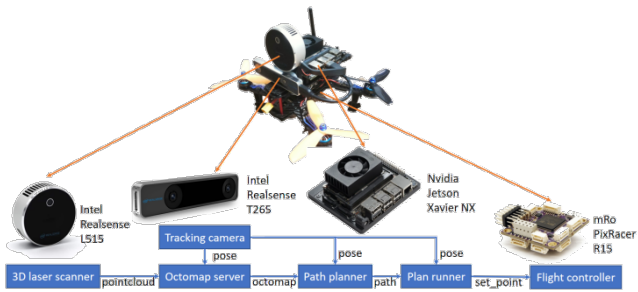

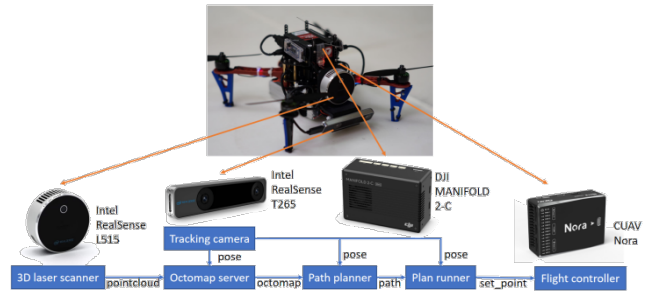

Figure 3. UAV prototype version one          Figure 4. UAV prototype version two

### 3.3.1 UAV Prototype Version One

After making comprehensive field tests and improvements, expect from the components we design by ourselves using carbon fiber laser cutting or 3D printing such as landing gear, camera mount, etc. The final stable version for UAV prototype version one uses the following components. For the base frame, we use Readytosky 250mm which is the smallest frame among all the current UAV prototypes that can complete the exploration tasks. For the onboard computer, we use Nvidia Jetson Xavier NX which is famous for its extendibility and lightweight. For the sensors, we mount two cameras for different purposes. One is Intel RealSense L515 which is used for generating point cloud information and constructed maps. The other is Intel RealSense T265 which is an RGBD camera used for providing localization information to the onboard computer for extended Kalman filter fusion. The controller we use for this UAV prototype is mRo PixRacer R15. The reason for this choice is that the PixRacer controller embeds the compass which can also provide localization information to the onboard computer for extended Kalman filter fusion. Two different localization sources will make the localization output more accurate. We utilize the PX4 and QGroundControl software compatible with the PixRacer controller to do calibration for the motors, get battery information for the drone, and communicate with the laptop. The controller can take the target position in three-dimensional space from our high-efficiency path planner as the input and output a series of electric signals to reach the target position at a certain velocity. To transfer the electric signals from the controller to the actual rotating speed of the motors, an appropriate electronic speed controller for each motor is crucial, an incompatible electronic speed controller will risk being breakdown by the high voltage signal and then generating inaccurate rotating speed which will lead to the unstable performance of the UAV. Thus, after testing, we find electronic speed controllers with the following tech specs can work stably on our UAV prototype. It should be compatible with 2S-6S Lipo, the maximum voltage tolerance should be at least 35A, the sensorless brushless motor electronic speed control firmware should be BLHeli_S, and the digital shot (Dshot) used for communication should support 150/300/600 levels. Based on the estimated time needed for completing the whole exploration process, we select the 6000mAh 4S Lipo battery that can enable a ~8min working time after one time fully charge. After finalizing these main components of the UAV, the total weight of the drone can be estimated (~1200g). Then we can find appropriate motors and propellers pair to provide enough thrust for the UAV prototype. To guarantee safety during flight, the total thrust should be two times the UAV payload.

So, the thrust should be at least 2400g. After testing, we choose iFlight XING-E 2208 2450KV Brushless Motor 4S together with three-blade 6045 (propeller diameter is 6 inches and propeller pitch is 4.5 inches) propellers to provide a ~2500g total thrust.

### 3.3.2 UAV Prototype Version Two

As some parts are common between the UAV prototype version one and the UAV prototype version two, we will only address the differences between these two models. In UAV prototype version two, we use a new onboard computer DJI Manifold2 because it has more extensions, better stability, and more computation power. We also use a new controller, CUAV Nora. It has a similar PX4 software structure, but the general encapsulation is better which provides more necessary ports, and its embedded sensors are more reliable and robust. Additionally, we change the distribution of the components to make it more compact which is good for control and path planning. Based on the tech specs of L515 that the depth field of view is 70 degrees by 55 degrees and the RGB sensor field of view is 70 degrees by 43 degrees, as shown in Fig. 3 and Fig. 4, the positions of two cameras have been rearranged to catch more useful features in the environment which is helpful for localization and reconstruction. New designs of landing gear and propeller guards are used to make the prototype stronger than the original.

Mathematical specifications and components for UAV prototypes:

UAV Prototype Components:

$$\text{Components: } C=\{C_1, C_2, \ldots, C_k\} \qquad \text{[Formula 7]}$$

Components $(C = \{C_1, C_2, \ldots, C_k\})$ include Nvidia Jetson Xavier NX for computational tasks and Intel RealSense cameras $(C_1, C_2)$ for environment perception.

Weight Calculation:

$$\text{Total Weight: } W_{total} = \sum_{i=1}^{k} w_i \qquad \text{[Formula 8]}$$

Total weight $(W_{total})$ of the UAV prototype includes components $(w_i)$ such as motors and sensors.

Motor and Propeller Selection:

$$\text{Thrust Requirement: } T_{required} \geq 2 \times W_{payload} \qquad \text{[Formula 9]}$$

Motor and propeller selections ensure $(T_{required} \geq 2 \times W_{payload})$ to maintain adequate thrust for stable flight.

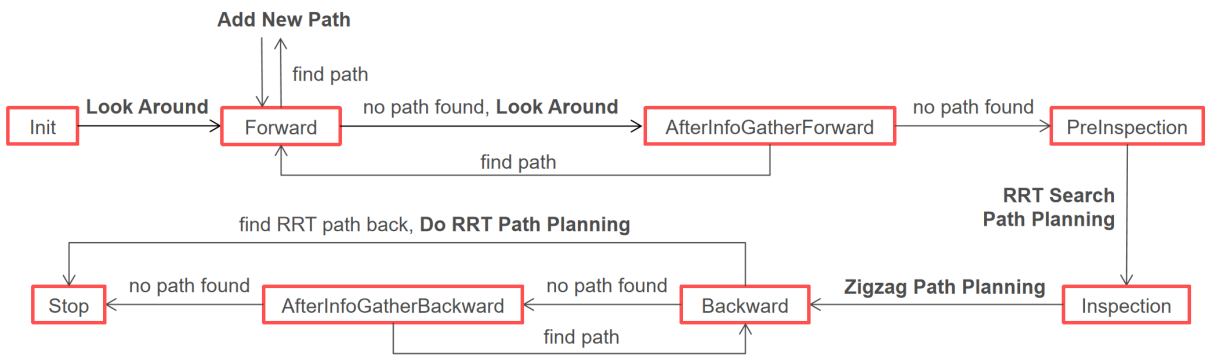

Figure 5. Path planner finite state transition logistic overview. The plain normal font means the conditional judgement, the bold font means the action taken, and the red box shows the current UAV state.

## 3.4 High-Efficiency Path Planner

To complete the tasks, we pre-define several finite states for the UAV, and Fig. 5 shows the basic state transition logistics of this high-efficiency path planner.

**Actions:** For the "look around" action, it is designed to gain more information from the environment to help the drone generate the path. Given the depth field of view and the RGB sensor field of view for the L515 camera, making the UAV rotates 90 degrees clockwise, then 180 degrees anti-clockwise, and then 90 degrees clockwise, will make the UAV gains more environmental information. It is especially useful at the early stage of the exploration when the feature in the constructed map is sparse and the environment with dense obstacles where there might be only a narrow feasible corridor for navigation. For the "add new path" action, it is designed to create a path segment for our controller to follow, in our specific algorithm implementation, the new path in this action is a move one meter straight forward path. The reason for choosing this path is that based on the attribute of the L515 camera, the camera can guarantee a high accuracy within three meters distance and moving straight is the most efficient action when there is no obstacle on the way, and we want to navigate to the dead-end of the tunnel as fast as possible. For the "RRT search path planning" action, it is designed to follow the Rapidly-exploring Random Trees algorithm [19].

The RRT search path planning algorithm is a sample-based search algorithm that is especially efficient in high-dimension path planning problems. In this problem, there are three dimensions in state space, the x, y, and z positions in the tunnel construction space. However, one drawback of the RRT algorithm is that it can not guarantee the optimality of the path and the generated path could be tortuous. We propose two optimization methods to remove redundant waypoints in the generated path. The first method is to iteratively (the max iteration number in our algorithm is 1000) and randomly pick two waypoints, try to connect them directly, and check if this new path will lead to a collision, if not, then we remove all waypoints in the original path between these two picked points. After applying the first method and having an optimized trajectory consisting of many straight-line path segments, our second optimization method is to randomly pick points on these path segments and try to connect them to build a collision-free new path, if it is collision-free, then redundant waypoints can be furtherly removed. After applying these two optimization methods, the generated path in the environments where the obstacles are sparse is largely simplified. For the "zigzag path planning" action, this action is designed to help perform 3D reconstruction of the dead-end of the tunnel. Based on the shape and the size of the tunnel, a zigzag path can be calculated to fully scan the entire tunnel end.

**States:** In the "Init" state, the drone takes off at the start position, hovering at one-meter height, and looks around to gain basic environment information and construct a map from empty. In the "Forward" state, the drone directly keeps going forward until there is no path found. In the "AfterInfoGatherForward" state, the drone gains more information after looking around, if the supplemental information helps to find a path to move forward, then back to the "Forward" state, else goes to the "PreInspection" state. In the "PreInspection" state, the drone needs to use the current constructed map and RRT algorithm to find a feasible collision-free path to navigate to the tunnel end. In the "Inspection" state, the drone performs zigzag path planning to help make a 3D reconstruction of the dead-end of the tunnel. Then in the "Backward" state, after the drone finishes the exploration, navigation, and 3D reconstruction tasks, it will try to find a direct straight backward path to fly back to its start point and the state will keep changing between the "Backward" state and the "AfterInfoGatherBackward" state. If no straight backward path is found, we will perform RRT path planning in which we set the start point as the target point. Based on our built map during the forward exploration process and following that generated RRT path, we can change to the "Stop" state. In addition, no path found in the "AfterInfoGatherBackward" state will also lead to the "Stop" state. In the "Stop" state, the drone will hover for a while and then land, this is the last state of our experiment.

Detailed description of path planning algorithm states and actions:

State Transition Logic:

$$\begin{bmatrix} s_{11} & s_{12} & \cdots & s_{1n} \\ s_{21} & s_{22} & \cdots & s_{2n} \\ \vdots & \vdots & \ddots & \vdots \\ s_{n1} & s_{n2} & \cdots & s_{nn} \end{bmatrix}$$ [Formula 10]

The state transition matrix（S）governs transitions between states（$s_{ij}$）during path planning.

Actions and Path Planning Methods:

Action Set: $A=\{A_1, A_2, \ldots, A_m\}$ [Formula 11]

Actions（$A=\{A_1, A_2, \ldots, A_m\}$）include path planning methods like RRT and zigzag for exploration and reconstruction tasks.

Optimization Criteria:

Optimization Function: $\min(\sum_{i=1}^{n} c_i x_i)$ [Formula 12]

Optimization criteria（$\min(\sum_{i=1}^{n} c_i x_i)$）ensure efficient path generation with minimal waypoints（$x_i$）satisfying constraints（$c_i$）.

# 4. Results and Discussion

## 4.1 General Experiments Setup

To evaluate the proposed prototypes' performance, we conduct simulation experiments, real-world wind test experiments, and real-world exploration and navigation experiments. The algorithm is implemented in ROS with C++ running on Intel Core i7-10750H CPU@2.6GHz for simulation experiments, and Nvidia Jetson Xavier NX for UAV prototype version one and DJI Manifold 2-C for UAV prototype version two for real-world tests. Overall, the entire system can run in real-time for the laptop and the UAV's onboard computer. The planner in the proposed prototypes can run up 30Hz by the UAV's onboard computer to guarantee an obstacle avoidance ability for low-speed dynamic obstacles. The Octomap [20] is used to represent the constructed map.

**4.2 Simulation Experiments**

To evaluate the effectiveness and robustness of the safe trajectory planning software framework, we prepare two different simulation environments[21].

**PX4 Gazebo Simulation Experiments Setup:** As shown in Fig. 6, to test the performance of the proposed planner, we build a PX4 Gazebo tunnel environment. The people in the tunnel are designed to move following the pre-defined linear trajectory[22]. One advantage of using the PX4 Gazebo simulation is that it is compatible with our current PX4 and Robot Operating System framework which makes it convenient to visualize the constructed map and generated path from the planner[23]. Thus, this simulation method is highly preferred in our experiments and recommended for any ROS-based planner[24]..

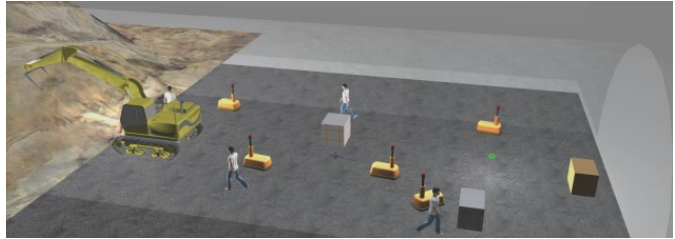

Figure 6. PX4 Gazebo simulation environments with walking people and moving robots.

**Airsim Unreal Engine 4 Simulation Experiments Setup:** As shown in Fig. 7, we build a customized tunnel environment in the Unreal Engine 4 environment with the Airsim plugin from Microsoft to test the performance of our proposed planner. One advantage of using Airsim Unreal Engine 4 simulation environment is that it can easily import the real-size 3D model built by us and the output visions from the camera can be visualized to give an intuitive idea about how a planner works in different stages[25].

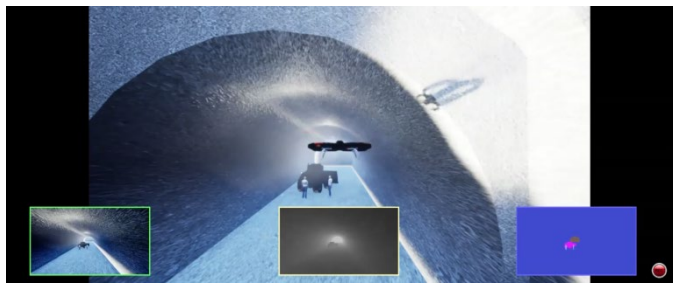

Figure 7. Airsim Unreal Engine 4 simulation environments with moving people and static forklift.

In this figure, the drone starts to do the navigation to the end. Three windows at the bottom show the RGD image, depth image, semantic segmentation result from the camera.

**Analysis and Discussion:** To analyze the performance of our proposed high-efficiency path planner, Fig. 8 shows the generated zigzag path from the planner, at this state, the UAV [26] has successfully navigated to the tunnel end while avoiding the moving workers at the same time and built the voxel map for backward navigation As the ROS visualization [27] is not compatible with Unreal Engine 4, Fig. 9 briefly shows the navigation process in simulation. In conclusion, after setting up two different simulation environments to verify the performance of the planner, results from both environment shows the ability of our proposed planner to finish the navigation and exploration tasks.

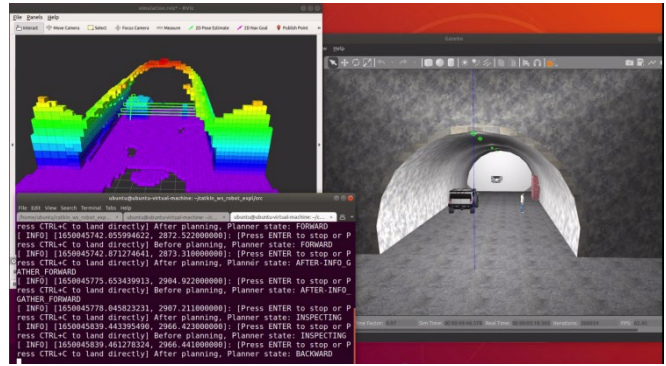

Figure 8. PX4 Gazebo navigation and exploration result. The left upper figure is the constructed voxel map, and the right figure is the actual performance in simulation.

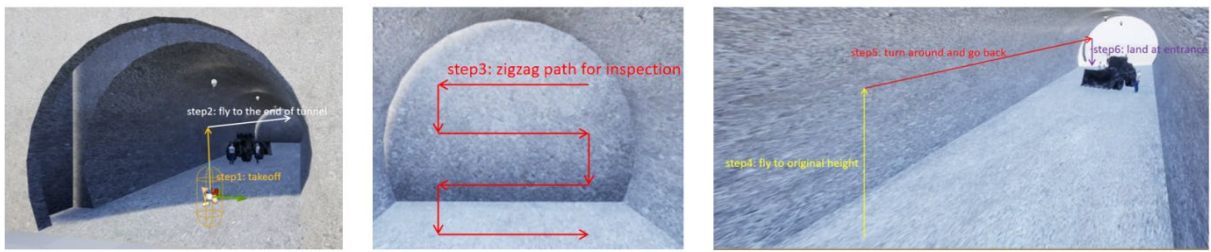

Figure 9. Airsim Unreal Engine 4 navigation result.

## 4.3 Real-World Experiments

To better evaluate the real-world performance of the UAV prototypes, we conducted exploration and navigation tests in our test field and the mock tunnel. The supplemental wind tests that simulate the tunnel wind disturbance are designed and performed to test the robustness of the UAVs' operation stability.

**Real-World Wind Test Experiments Setup:** The real-world wind experiments are conducted in Mill 19, Carnegie Mellon University. The OptiTrack motion capture system is used to collect the data and quantitatively show the corresponding drifts given different levels of wind and drone actions. The top view of the wind test experiment setup is shown in Fig. 10.

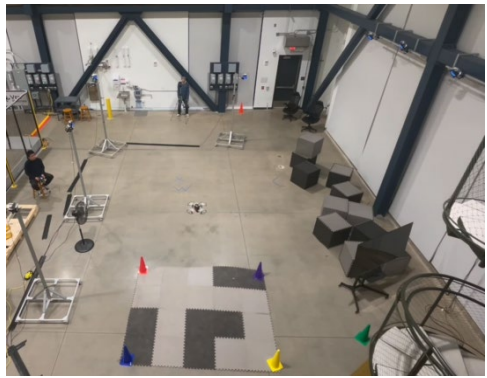

Figure 10. top view of the wind test experiment setup.

**Real-World Exploration and Navigation Experiments Setup:** The real-world exploration and navigation experiments consist of three warm-up experiments to guarantee the status of the sensors works fine and one environment exploration test. The experiments are conducted in several different test fields. For warm-up experiments, firstly we conduct a manual mode flight test to control the drone with only the manual controller to test the functionalities of the UAV hardware platform are normal. Secondly, we conduct an Octomap mapping test to make sure that the L515 camera can correctly utilize the collected environment information to build the map. In some cases, due to the attribute of the L515 camera, environments with very sparse features or with special materials that are highly reflective or can absorb most laser sent from the L515 camera will lead to the failure of building map accurately which is dangerous for actual flight test. Then, the last warmup experiment is the position mode flight test. Given several target points, this experiment is to test the controller and the UAV's ability to track target points in the real world. The real-world experiment is performed in many locations such as Baker Hall, Carnegie Mellon University, and the mockup tunnel at Fukushima, Japan as shown in Fig. 11 to verify the performance.

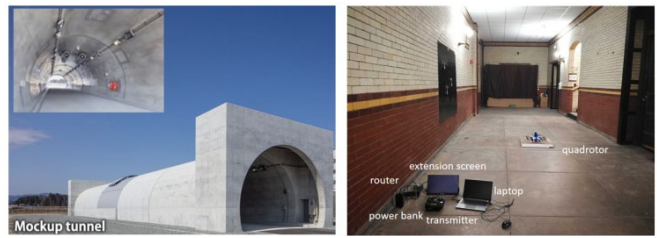

Figure 11. Mockup tunnel at Fukushima, Japan (left). Baker Hall, Carnegie Mellon University (right).

**Analysis and Discussion:** To analyze the control stability of the designed autonomous UAV prototypes,the drift results of the wind test experiments under three different wind levels where the speeds of the wind are measured with the anemometer in the hovering state are shown in Tab. 1, and in going straight state is shown in Tab. 2.

Table 1. Hovering Wind Test.

| Wind levels | Max drift in x-axis | Max drift in y-axis |
| --- | --- | --- |
| No wind | 0.189m | 0.288m |
| Low (2.30m/s) | 0.381m | 0.294m |
| Middle (2.71m/s) | 0.471m | 0.283m |
| High (3.24m/s) | 0.535m | 0.213m |

Table 2. Going Straight Wind Test.

| Wind levels | Intrinsic drift in y-axis | Max drift in y-axis |
| --- | --- | --- |
| No wind | 0.200m | 0.460m |
| Low (1.86m/s) | 0.062m | 0.448m |
| Middle (2.36m/s) | 0.013m | 0.333m |
| High (2.71m/s) | 0.338m | 0.625m |

Based on the drift results in the hovering wind test, as we set the wind blowing along the x-axis, the drifts in the x-axis are larger with the increase of the wind levels while the drifts in the y-axis are relatively close. Considering the intrinsic tracking error from the controller, we can conclude that with the levels of the wind increase, the drone does have more oscillation behaviors (in Fig. 12). But in general, under all wind levels, compared to the no wind hovering performance, the drone can keep a quite stable behavior. As we set the wind blows along the y-axis, the intrinsic drifts (the drifts between the take-off position and land position) in the y-axis and the max drifts in the y-axis are considered and compared. Based on the drift results in going straight wind test, we can conclude that (in Fig. 13) while the controller is tracking the trajectory given a target position, a higher wind level will lead to more oscillations of the original trajectory and the position where the UAV experiences the wind will occur a drift peak off the original trajectory. However, considering the drift peak, the error compensation ability of the controller can make the general drift acceptable and reach the target position at last.

To analyze the comprehensive performance of our drone in the real world. The constructed map result after exploration and navigation experiment is shown in Fig. 14. Based on our experience gained from the real-world experiments, the performance of the drone can be influenced by many external factors, such as the quality of the communication connections, the battery remaining power, and the materials of the environments. Overall, from the experiments, we can conclude that our proposed UAV prototypes can finish the exploration and navigation tasks, while prototype version two has a more stable behavior due to the improvement of the hardware performance and distribution.

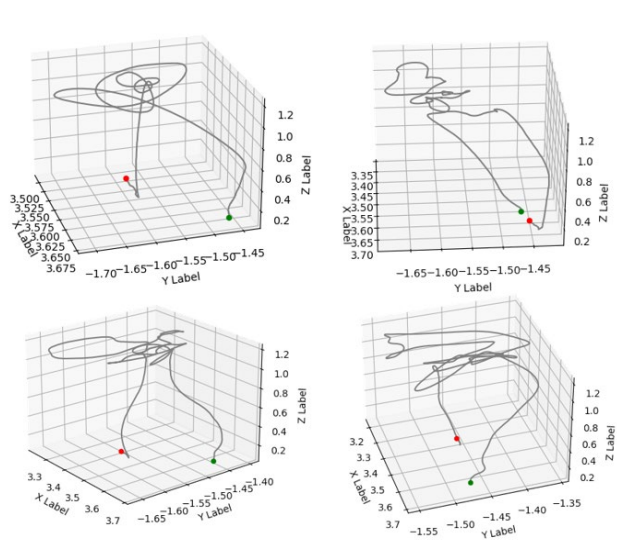

Figure 12. Trajectory visualization in hovering wind test. side view, take off position (green dot), land position (red dot). No wind (upper left), Low (upper right), Middle (lower left), High (lower right).

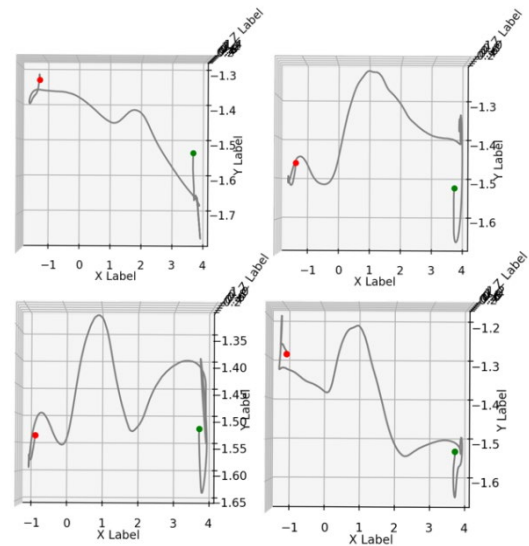

Figure 13. Trajectory Visualization in going straight wind test. top view, take off position (green dot), land position (red dot). No wind (upper left), Low (upper right), Middle (lower left), High (lower right).

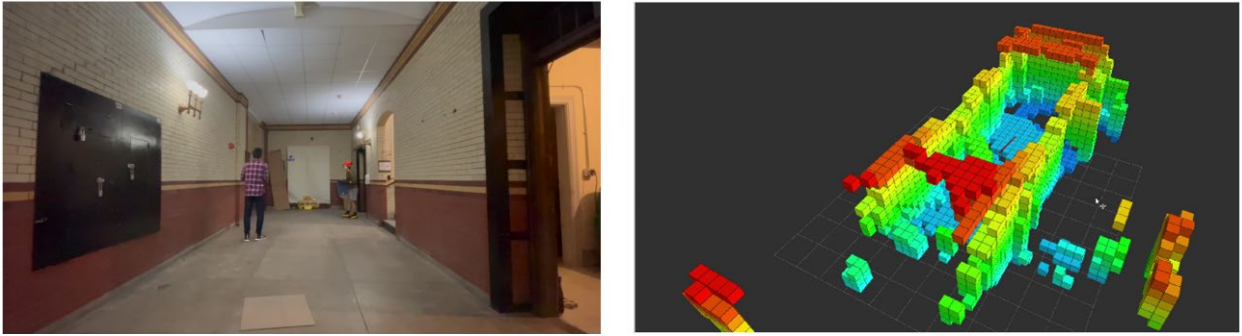

Figure 14. Real-world exploration and navigation experiments.
The UAV (in middle of the fig) performs navigation with moving people obstacles. (left),
Constructed voxel map after exploration. (right).

## 5. Summary and Conclusions

In conclusion, this paper has presented novel designs of autonomous UAV prototypes tailored for inspecting GPS-denied tunnel construction environments, featuring dynamic obstacles such as humans and robots. Our research focused on developing robust UAV hardware platforms integrated with advanced sensor technologies and efficient motion planning algorithms. Through simulation and real-world experiments, we demonstrated the effectiveness of our prototypes in navigating and exploring complex environments autonomously.

Looking forward, the prospects of this research lie in several promising directions. Firstly, further refinement of our UAV prototypes to enhance their stability and adaptability in varied environmental conditions remains a priority. Additionally, incorporating machine learning techniques [28-68] for real-time decision-making and obstacle recognition could significantly improve operational efficiency and safety. Moreover, exploring cooperative navigation strategies among multiple UAVs could extend the application scope to larger-scale infrastructure inspections.

In summary, this study contributes to advancing UAV technology's capabilities in challenging real-world settings and lays the groundwork for future developments in autonomous aerial inspection systems.